\documentclass[conference]{IEEEtran}

\usepackage[letterpaper, left=1in, right=1in, bottom=1in, top=0.75in]{geometry}
\usepackage{graphicx}
\usepackage{amsmath}
\usepackage{array}
\usepackage{epsfig}
\usepackage{amssymb}
\usepackage{color}

\begin{document}
\title{A Survey of Conventional and Artificial Intelligence / Learning based Resource Allocation and Interference Mitigation Schemes in D2D Enabled Networks}

%\author{\IEEEauthorblockN{Kamran Zia}
%\IEEEauthorblockA{National University\\of Sciences and
%Technology\\
%Email: kamran.zia@cae.nust.edu.pk}
%\and
%\IEEEauthorblockN{Nauman Javed}
%\IEEEauthorblockA{National University\\of Sciences and
%	Technology\\
%	Email: nauman.javed@cae.nust.edu.pk}
%\and
%\IEEEauthorblockN{Nadeem Sial}
%\IEEEauthorblockA{Department of Informatics\\Kings College London\\
%	Email: muhammad.sial@kcl.ac.uk}
%\and
%\IEEEauthorblockN{Sohail Ahmed}
%\IEEEauthorblockA{Air University Islamabad\\
%	Pakistan\\
%	Email: sohailahmed71@gmail.com}
%\and
%\IEEEauthorblockN{Hifsa Iram}
%\IEEEauthorblockA{National University of Computer\\ and Emerging Sciences\\
%	Email: hifsairam@gmail.com}}
\author{\IEEEauthorblockN{Kamran Zia$^{1}$, Nauman Javed$^1$, Muhammad Nadeem Sial$^2$, Sohail Ahmed$^3$\\ Hifsa Iram$^4$, Asad Amir Pirzada$^1$}
	\IEEEauthorblockA{$^1$National University of Sciences and Technology,Islamabad, Pakistan, $^2$ Kings College London \\ $^3$Air University Islamabad, Pakistan, $4$ National University of Computer and Emerging Sciences\\
		\footnotesize{kamran.zia@cae.nust.edu.pk,nauman.javed@cae.nust.edu.pk, muhammad.sial@kcl.ac.uk,sohailahmed71@gmail.com}\\
		\footnotesize{hifsairam@gmail.com, pirzada@cae.nust.edu.pk}}}

%Telephone: (800) 555--1212, Fax: (888) 555--1212}


\maketitle

\begin{abstract}
5th generation networks are envisioned to provide seamless and ubiquitous connection to 1000-fold more devices and is believed to provide ultra-low latency and higher data rates up to tens of Gbps. Different technologies enabling these requirements are being developed including mmWave communications, Massive MIMO and beamforming, Device to Device (D2D) communications and Heterogeneous Networks. D2D communication is a promising technology to enable applications requiring high bandwidth such as online streaming and online gaming etc. It can also provide ultra- low latencies required for applications like vehicle to vehicle communication for autonomous driving. D2D communication can provide higher data rates with high energy efficiency and spectral efficiency compared to conventional communication. The performance benefits of D2D communication can be best achieved when D2D users reuses the spectrum being utilized by the conventional cellular users. This spectrum sharing in a multi-tier heterogeneous network will introduce complex interference among D2D users and cellular users which needs to be resolved. Motivated by limited number of surveys for interference mitigation and resource allocation in D2D enabled heterogeneous networks, we have surveyed different conventional and artificial intelligence based interference mitigation and resource allocation schemes developed in recent years. Our contribution lies in the analysis of conventional interference mitigation techniques and their shortcomings. Finally, the strengths of AI based techniques are determined and open research challenges deduced from the recent research are presented.
\end{abstract}


% \fi
\IEEEpeerreviewmaketitle

\section{Introduction}
The ever-increasing demands of mobile users is the main reason for further expansion of the network in terms of capacity and throughput. According to the Ericsson Mobility Report ~\cite{ericsson2015ericsson}, mobile subscription will rise from 7.3 to 9 billion users worldwide in 5 years and per user data usage will rise form 1.4 GBs to 8.9 GBs. Similarly, Cisco Visual Networking Index ~\cite{index2013global} indicated that the total mobile traffic grew 81 percent and connection speeds doubled in the year 2013. Moreover, applications like internet of things (IoT), machine to machine communication, personalized TVs, Video streaming and video conference calls and self-driven cars require network with high bandwidth, data rate and latencies for their operation. The number of users requiring connection has gone manifold and network operators are facing resource shortage to provide services to such large number of users. The major requirements laid down by Next Generation Mobile Network (NGMN) alliance ~\cite{alliance20155g} for the 5th generation networks include data rates up to tens of Gbps, latencies as low as 1 ms, 1000-fold more connected devices and 10 times more battery efficiency. Different technologies are being developed and researched for meting these requirements and they include Massive MIMO and beamforming, full duplex communications, mmWave communications, UL-DL decoupled access, separation architecture for control plane and data plane, Device to Device communication and multitier heterogeneous networks with multiple radio access technologies (e.g. LTE, CDMA-2000, HSPA and HSPA+, WiFi etc) ~\cite{dahlman20145g}. Out of these, Device to Device communication (D2D) is a technology in which two closely spaced devices communicate directly with each other without relaying the data through base stations. Due to the increase in the multimedia and online gaming applications, the bandwidth requirement of users has increased and D2D communication is a promising technology to meet these application requirements. Moreover, D2D communication can provide higher network spectral efficiency by re-using the cellular frequency resources due to its short distance communication. Owing to the performance benefits of D2D communication, standardization agencies like 3rd Generation Partnership Program (3GPP) has recognized the importance of D2D communication for future mobile networks and laid down criteria and performance requirements for proximity-based services in its release 12 ~\cite{lin2014overview}.
\begin{figure*}
	\centering
	\includegraphics[width=1\linewidth]{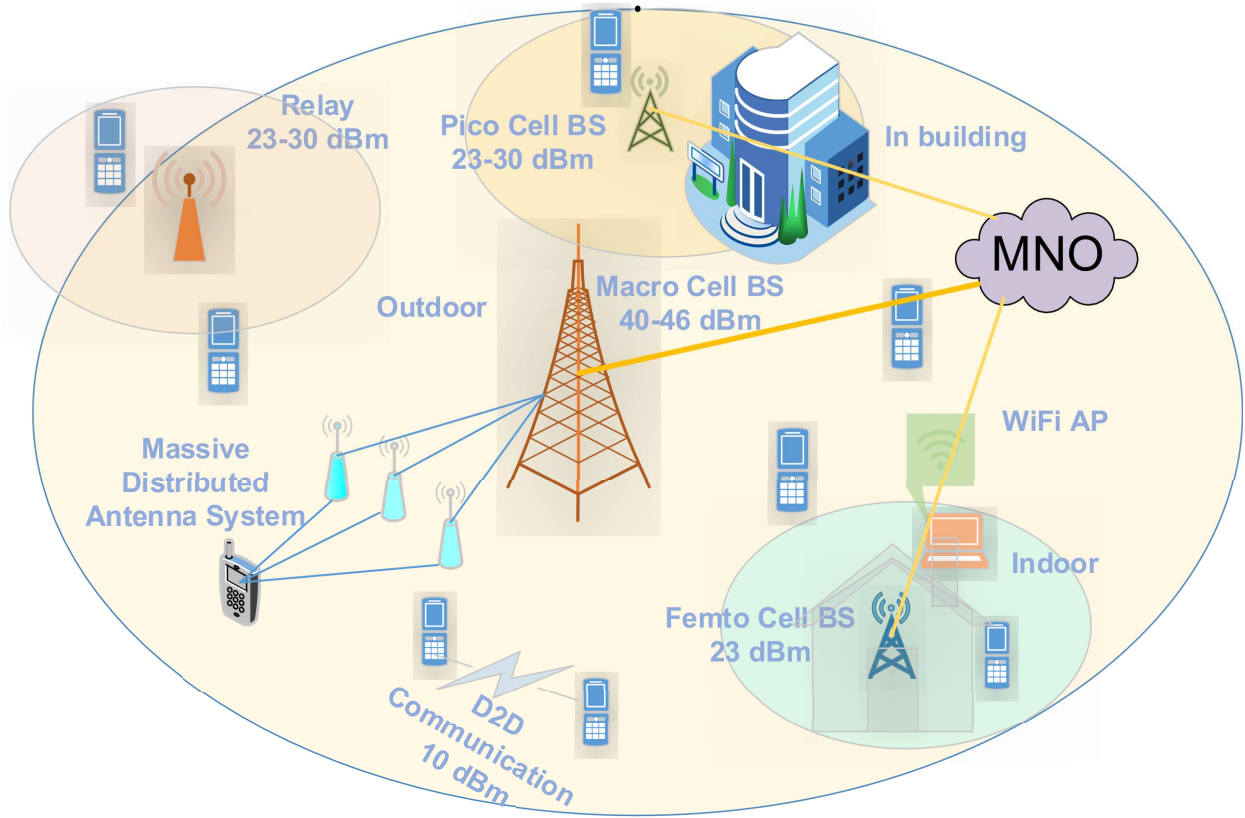}
	\caption{Infrastructure of 5G Heterogeneous Network}
	\label{fig:Hetnet2}
\end{figure*}

5th generation networks will have multiple radio access technologies (RAT) and multiple cells with different cell radius and transmit powers thus making it a heterogeneous network.  The next generation radio access network (RAN) system which is also called as the 5G RAN system will be combination of evolved and revolved multiple cooperating radio access technologies (RATs). Therefore, the architecture of the 5G RAN system will constitute evolved versions of 2G (GPRS/EDGE), 3G (HSPA/UMTS), 4G (LTE-Advanced), WLAN (WiFi) and machine-type communications (MTCs) ~\cite{olwal2016survey}.  Furthermore, due to higher capacity requirement of future networks, 5G RAN comprises of multiple tiers of heterogeneous networks (HetNets). The radio access technology of each base station with asymmetrical transmit powers will determine the cell sizes and will add to the complex interference scenario in the uplink and downlink of the network. The architecture of the 5G RAN systems will consist of macro cell and small cells (i.e., microcell, picocell, femtocell, relay) and device to device (D2D) based communication ~\cite{agrawal2009heterogeneous}. Usually the macro base stations serve the macro cells and have high transmit powers ranging from 43 dBm to 46 dBm and antenna gains close to 10 to 15 dBi ~\cite{lu2015wireless}. They are suitable for the remote and rural areas and have bigger coverage areas. Micro and pico cells are served by the micro an pico base stations with transmit powers ranging from 23 to 30 dBm and antenna gain of 0 to 5 dBi and they are suitable for short range urban areas.  Similarly, Femto base stations are the user deployed base stations with transmit powers less than 23 dBm for increasing the coverage of the network. Due to this complex multitier network structure with different RAT, the resource allocation techniques employed in 4G RAN are no longer feasible therefore research community is working on new techniques including machine learning especially Artificial Intelligence (AI) based techniques to enable the high capacity and high throughput requirements of the future 5G networks. The infrastructure of a typical heterogeneous 5G network is shown in Figure \ref{fig:Hetnet2}.

D2D communication in this multitier HetNet adds to another tier thus making interference mitigation and resource allocation more complexer. D2D communication can takes place in either dedicated mode where D2D pairs uses dedicated frequency resources to communicate with each other ~\cite{mach2015band}. In this mode the interference caused to the cellular user is under control because of orthogonal frequencies being used by the D2D pair however this use of D2D communication is not spectral efficient. The user densification has given rise to the scarcity of the spectrum in order to meet Quality of Service (QoS) requirements of the users therefore D2D pairs are required to use the frequency which is already being used by the cellular users. This mode of D2D communication is known as underlay mode and it is more spectral efficient but the challenge comes in the allocation of resources to these pairs such that they do not cause significant interference to the cellular user. Researchers have already identified the problem complexity and have been developing different techniques to control interference and allocate resources in D2D enabled heterogeneous networks.
\begin{table}[h!]
	\begin{center}
        \scriptsize
		\caption{Abbreviations}
		\label{tab:Abbr}
        \begin{tabular}{|c|l|}
  \hline
  \textbf{Abbreviation} & \textbf{Definition} \\
  \hline
  RAN & Radio Access Network \\
  \hline
  RAT & Radio Access Technology \\
  \hline
  RRM & Radio Resource Management \\
  \hline
  3GPP & $3^{rd}$ Generation Partnership Program \\
  \hline
  NGMN & Next Generation Mobile Networks \\
  \hline
  IMT & International Mobile Telecommunications \\
  \hline
  MTC & Machine type Communication \\
  \hline
  UL & Uplink \\
  \hline
  DL & Downlink \\
  \hline
  CDMA & Code Division Multiple Access \\
  \hline
  HSPA & High Speed Packet Access \\
  \hline
  LTE & Long Term Evolution \\
  \hline
  GPRS & General Packet Radio Service \\
  \hline
  UMTS & Universal Mobile Telecommunication System \\
  \hline
  EDGE & Enhanced Data for GSM Evolution \\
  \hline
  MIMO & Multiple Input Multiple Output \\
  \hline
  CSI & Channel State Information \\
  \hline
  UE & User Equipment \\
  \hline
  DUE & D2D User Equipment \\
  \hline
  CUE & Cellular User Equipment \\
  \hline
  MME & Mobility Management Entity \\
  \hline
  P-GW & Packet Data Gateway \\
  \hline
  TDD & Time Division Duplexing \\
  \hline
  OFDM & Orthogonal Frequency Division Duplexing \\
  \hline
  MINLP & Mixed Integer Non-Linear Programming \\
  \hline
  TTI & Transmission Time Interval \\
  \hline
  eMBB & Enhanced Mobile Broadband \\
  \hline
  SPM & Service Provision Management \\
  \hline
  URLLC & Ultra-Reliable Low Latency Communication \\
  \hline
  QoS & Quality of Service \\
  \hline
  MNO & Mobile Network Operator\\
  \hline
    \end{tabular}
	\end{center}
\end{table}

A number of surveys related to D2D communication in heterogeneous network have been done ~\cite{mach2015band}, ~\cite{liu2015device} and network architecture supporting D2D communication, D2D communication scenarios, interference mitigation techniques and research directions have been presented in these surveys. Surveys related to resource allocation schemes for 5G Hetnets have been done ~\cite{olwal2016survey} and conventional resource allocation schemes have been discussed but no survey has been done comparing Conventional Resource Allocation techniques and Artificial Intelligence / Machine Learning based techniques.  Some surveys ~\cite{jiang2017machine} ~\cite{wang2015artificial} identifying the importance of Machine Learning and Artificial Intelligence in 5G networks have been presented but they are not focused on problem of resource allocation and interference mitigation. In this paper we survey conventional interference mitigation and resource allocation techniques which have been proposed in the past few years and highlight the shortcoming in these techniques. We then survey artificial intelligence-based techniques developed in recent years for interference mitigation and resource allocation to illustrate the requirement of AI in future networks. In the end, future research directions to fulfill 5th generation network requirements as laid down by 3GPP and NGMN have also been proposed. The rest of the paper is organized as follows. Section II describes the details about D2D communication management, communication scenarios and different D2D communication modes. Section III contains the discussion on conventional resource allocation and interference mitigation techniques for D2D enabled network. Section IV contains artificial intelligence and learning based techniques for interference mitigation and resource allocation. Section V presents the challebges and future research directions in the field and section VI concludes this paper.

\section{Classification of D2D Communication}
D2D communication can be classified according to the several distinctive categories like D2D management, D2D scenarios and D2D communication mode according to use of radio resource.
\begin{figure}
	\centering
	\includegraphics[width=1\linewidth]{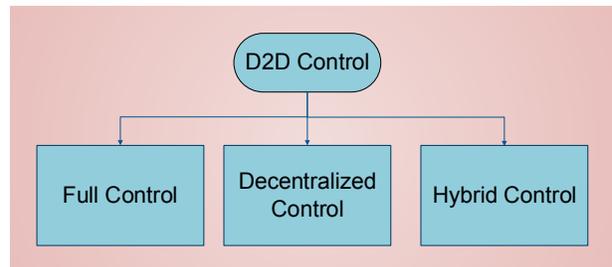}
	\caption{D2D Management}
	\label{fig:D2DControl}
\end{figure}

\subsection{D2D Management}
The network’s involvement in control of initiation of D2D communication defines its classification from management point of view. (See Figure \ref{fig:D2DControl})

\subsubsection{Full Control}	
 The level of control can be either full control where the network is in full control of D2D peers discovery and connection initiation, After the discovery and connection establishment, the network allocates power and radio resource to the D2D peers for communication with each other. The advantage of full control is that the network can easily coordinate and allocate power and radio resource to the users. Thus, in this way the harmful interference between D2D and cellular users can easily be avoided. Moreover, the base station can easily prioritize the individual transmissions to meet the QoS requirements of the users. However, with additional management tasks in hand by the base stations for D2D control, significant overhead is added to its processing. Base stations also have to know the channel state information (CSI) and share it with other base stations and this exchange is very demanding in terms of signaling ~\cite{tao2012qos}.

 \subsubsection{Decentralized Control}
 The D2D control can also be decentralized where D2D peers communicate with each other autonomously with very less intervention by the base stations ~\cite{lei2012operator}. The network or base station is only responsible for authentication of the devices during the connection setup between the devices. Afterwards, transmission power selection and radio resource selection are autonomously done by the devices themselves. Most of the functions are solely controlled by the D2D devices (DUEs) in this distributively controlled D2D communication. The disadvantage in this mode is that the DUEs cause significant interference to the conventional cellular users (CUEs) and their QoS parameters cannot be met by the network. Significant intelligence is required to be added to the DUEs to select the radio resource. Interference control techniques are also needed to be developed for meeting the QoS requirement of all the users of the network. Another solution to this approach is to use unlicensed frequency bands which are being used by WiFi and bluetooth based devices however use of such bands will cause interference for DUEs as there is no control of network over these unlicensed bands and devices operating in this band.

 \subsubsection{Hybrid  Control}
Another classification of D2D communication on the basis of D2D management and control is the hybrid control mode ~\cite{chen2010time}. In this mode, network (base station) controls the authentication, connection establishment and resource allocation to the DUEs while DUEs themselves can also select transmission power levels and radio resource in a decentralized manner based on the measurement of the channel state. Hybrid mode offers a good tradeoff to the network operators in terms of reduced signaling overhead and control of DUEs. It also maintains the QoS requirement of CUEs as well as DUEs in the network.

\subsection{D2D Discovery}
D2D discovery is another main aspect of D2D communication in a network and it significantly affects the interference caused to the CUEs. The purpose of this discovery process is to find the potential users that can communicate directly with each other to increase network capacity and benefit from the close distance communication in terms of low latency and higher throughput. The discovery process takes place in two stages namely the discovery initiation and discovery control.
\begin{figure}
	\centering
	\includegraphics[width=1\linewidth]{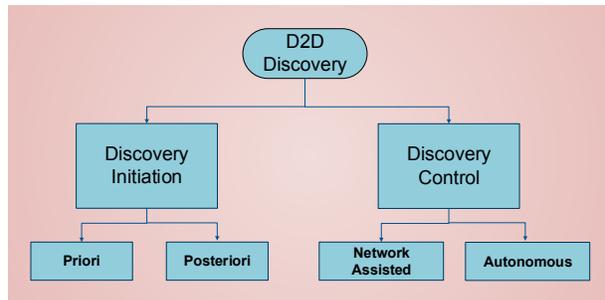}
	\caption{D2D Discovery}
	\label{fig:D2DDiscovery}
\end{figure}

The discovery initiation can be ‘priori’ or ‘posteriori’. In priori discovery, the DUEs do not communicate before the discovery is done and it is commonly done when two devices want to share some content with each other. In posteriori discovery, the devices communicate with each other after the discovery is done. (See Figure \ref{fig:D2DDiscovery})

The discovery process is also controlled with different levels of involvement by the network. The discovery process can either be network assisted or it may be fully autonomous where devices discover the closely located other cellular devices that can be potential D2D peers. The network assisted discovery is easier as the network is aware of the devices locations and channel state information however it adds to the signaling overhead to the base station. The autonomous discovery has the advantage of low signaling overhead but the discovery process itself can drain the battery of the DUEs. (See Figure \ref{fig:D2DDiscovery})

Different D2D peers discovery schemes have been presented in literature incluidng both network assisted and autonomous discoveries. A network assisted discovery procedure is presented in ~\cite{yang2013solving}. In this procedure, the packet data network gateway (P-GW) detects the potential D2D users and then a message exchange takes place between Mobility Management Entity (MME), Base Station and the UEs participating in the discovery process. After the establishment of D2D connection, direct communication takes place between the DUEs. This procedure however adds significant overhead to the processing tasks performed by the base stations therefore authors in ~\cite{nguyen2014network} presented a D2D discovery procedure with lesser overhead. The UEs performs discovery in time slot-based manner using frequency multiplexed discovery channels. During certain time intervals, the devices search and listen to the discovery signals from other devices and establish connections. The number of discoveries is significantly increased by this discovery procedure with less overhead to the base stations.

Qualcomm has developed an autonomous discovery technique namely FlashLinQ ~\cite{wu2013flashlinq}. FlashLinQ is a synchronous Time Division Duplexing (TDD) Orthogonal Frequency Division Multiplexing (OFDM) system and it is designed to enable the discovery of UEs autonomously and continuously for D2D communication at high data rates. It is designed over licensed band at 2.586 GHz carrier frequency and bandwidth of 5 MHz. A Dynamic Source Routing (DSR) based D2D peer discovery is introduced in ~\cite{kaufman2009interference} in which network broadcast discovery packets through flooding. The information in discovery packets include transmission power of the devices, channel number and measured CSI. The receivers measure the SINR and path losses based on this information and estimates their transmit powers to be heard by the transmitters. If a bidirectional link can be established, D2D connection is made between the devices for direct communication.

\begin{figure*}
	\centering
	\includegraphics[width=1\linewidth]{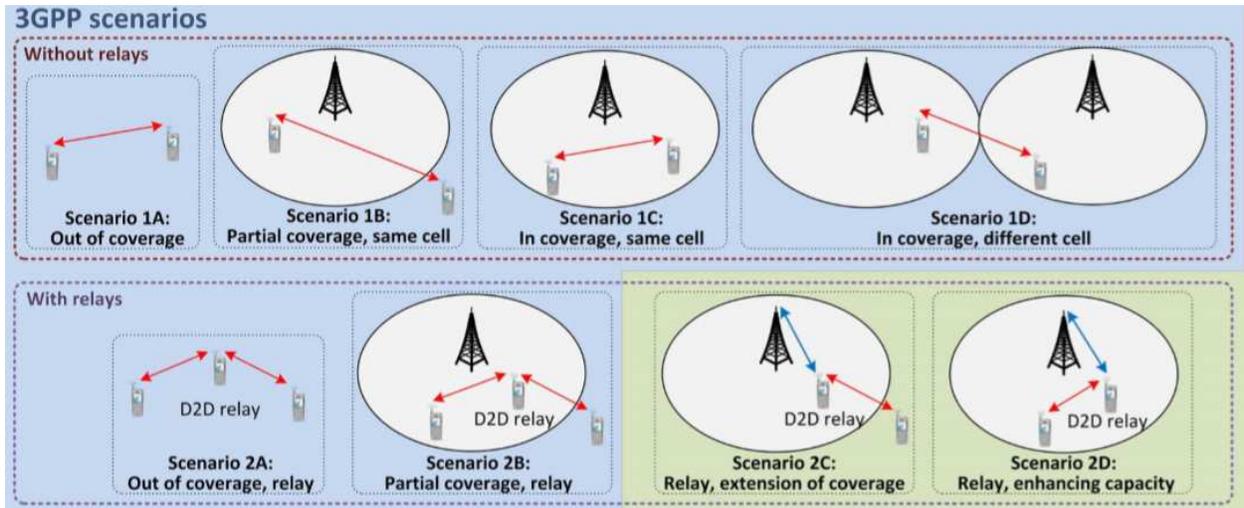}
	\caption{D2D Communication Scenarios laid down in 3GPP ~\cite{mach2015band}}
	\label{fig:D2DScenarios}
\end{figure*}

\subsection{D2D Scenarios}
In this section, different D2D communication scenarios (see Figure \ref{fig:D2DScenarios}) are discussed as follows:

\subsubsection{Coverage}
According to the coverage, the D2D communication can be categorized as follows
\subitem -In Coverage: Both the D2D peers are in coverage of the cellular network. (Scenario 1C)
\subitem -Partial Coverage: One of the users in D2D pair is out of coverage and one is in coverage of the cellular network. (Scenario 1B)
\subitem -Out of Coverage: Both the D2D peers are out of coverage of the cellular network (Scenario 1A). This scenario is considered in 3GPP for public safety cases when network is temporarily disabled due to some disaster like earth quake, floods etc.
\subsubsection{Type of D2D Communication}
This classification of D2D communication expresses how many DUEs are involved in the D2D communication:
\subitem -One-to-One Communication: One D2D communication pair communicating with each other.
\subitem -One-to-many Communication: One DUE is communicating with multiple DUEs simultaneously by broadcasting or multicasting the information.
\subsubsection{Area of D2D Communication}
This classification is based on whether same cell or different cells are serving the DUEs.
\subitem -Same Cell: The participating UEs are located in the same cell and are attached to same base station.
(Scenario 1C)
\subitem -Different Cell: The D2D peers are linked to different base stations and are located in different cells.
(Scenario 1D)
\subsubsection{Relaying Functionality}
If there is a requirement of retransmitting data, the DUEs can act as relay as well for multiple purposes.
\subitem -Enhance Capacity: The D2D pair communicating with each other is in coverage of base station. (Scenario 2D)
\subitem -Extend Coverage: The out of the coverage DUE can use other DUE to reach the base station. (Scenario 2C)

Depending on these classification, several scenarios can be defined as shown in figure \ref{fig:D2DScenarios}. These scenarios are defined by 3GPP standardization group in their release 12 for proximity-based services ~\cite{lin2014overview}. 3GPP has defined scenarios considering both the relay functionality as well as without relaying functionality.  The simplest scenario is the one where both DUEs lie in the same cell coverage.
\subsection{D2D Communication Modes}
D2D communication can take place in following two modes:
\subsubsection{Dedicated Mode}
The DUEs communicate directly with each without intervention of the base station in Dedicated Mode. However, the base station is still responsible to assign radio resource to the D2D pair for direct communication. The radio resources assigned to D2D pairs are orthogonal to the resources assigned to the CUEs therefore, there is no interference caused to the cellular users (CUEs) but the spectral efficiency of this mode is low because of dedicated resource usage. This mode of D2D communication is also known as overlay mode in literature. Advantage of this mode is that the base station does not need to implement interference mitigation techniques for meeting QoS requirements.
\subsubsection{Shared Mode}
In shared mode, the DUEs reuse the radio resources which are already being used by the CUEs and therefore there is strong interference caused by DUEs to the CUEs. This adds to the signaling overhead to the base station as it has to intelligently assign resources to the DUEs to avoid interference among DUEs and CUEs and meet their QoS requirements. Shared mode is also known as underlay mode or non-orthogonal mode in some literature. The spectral efficiency of SM is quite higher but it is quite complex from implementation point of view. The DUE can either use uplink (UL) radio resource or downlink (DL) radio resource however UL radio resources are usually preferred because the interference is caused to the base stations and transmission powers of UEs are quite lower than base stations therefore interference caused is also quite lower.

D2D communication can significantly increase the network capacity however it can be best achieved when spectrum is efficiently utilized and sophisticated intercell and intracell interference mitigation techniques are developed. Researchers have been developing different resource allocation and interference mitigation schemes for D2D enabled networks to fulfill the envisioned requirements of 5G network.  In the next section, we will be surveying the RRM and interference mitigation techniques developed by the communication engineers and researchers in the recent years to control the complex interference between DUEs and CUEs as well as intercell interference in single tier and multitier Hetnets.
\section{Conventional rrm and interference mitigation techniques}

D2D communication in underlay mode requires intelligent selection of radio resources to minimize interference to the CUEs. There can be different interference scenarios when the radio resources are shared among DUEs and CUEs. In one scenario, DUEs can cause interference to the CUEs and affect their SINR. Similarly, CUEs can also cause interference to the DUEs and affect their performance and there can be mutual interference between DUEs and CUEs.

\subsection {Interference Mitigation through Mode Selection}
\begin{figure}
	\centering
	\includegraphics[width=1\linewidth]{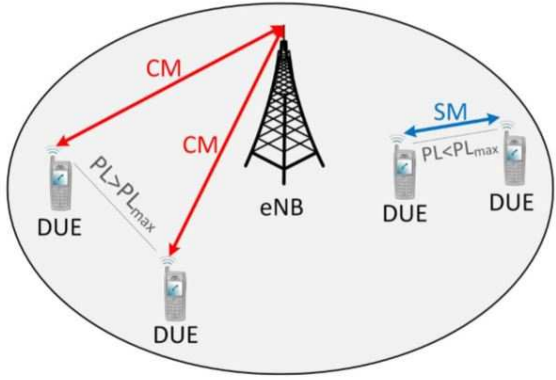}
	\caption{Mode selection based on path loss ~\cite{xing2010investigation}}
	\label{fig:PL}
\end{figure}

The interference control can be done by selecting the communication mode for D2D communication between the DUEs. As discussed earlier, there can be two communication modes for D2D communication namely dedicated mode and shared mode. Authors in ~\cite{xing2010investigation} considered a path loss model-based communication mode selection where the DUEs determines the path loss between them. If the path loss is greater than certain threshold path loss, D2D communication does not takes place and if its less than threshold, shared mode is selected. The basic principle of this selection is shown in Figure \ref{fig:PL}. Mode selection solely according to the path loss model is not optimal and interference control by this method is far from optimized solution.

Distance based mode selection scheme is presented in ~\cite{elsawy2014analytical}. The authors considered the distance between the base station and DUEs and mutual distance between the DUEs. The mode selection scheme accounts for both the cellular link quality as well as D2D users link quality. The mode selection is done if the D2D link quality is better than cellular user and distance of DUE from base station is greater than certain threshold. Authors have also included truncated channel inversion-based power control to control the interference between DUEs and CUEs. The author has proved that his mode selection scheme outperforms the mode selection scheme based on distance between DUEs only in terms of outage probability of CUEs. The mode selection between SM and DM is however not considered in this paper.

Authors in ~\cite{janis2009device} determined the benefit of D2D communication in terms of capacity enhancement if it is enabled and which mode is most appropriate to be selected. The sum rate maximization calculated according to the Shannon Capacity formula is taken as the criterion for mode selection. If CM mode gives higher sum rate, it is selected other SM or DM mode is selected. The selection of SM or DM is done according to the distance between the DUEs and base station. If the distance between base station and DUEs is greater i-e if D2D pair is far from base station, SM mode is selected in UL direction because interference to the base station is lower. If this distance is smaller, DM mode is selected for better efficiency. So far, the techniques discussed considered static channel conditions and does not account for the continuously varying channel conditions which is present in the practical networks.

For varying channel conditions, a dynamic mode selection scheme is required therefore authors in ~\cite{han2012subchannel} presented a partial solution where the network dynamically and opportunistically selects the mode for D2D communication. Dedicated mode and Conventional communication mode (CM) is considered in his study, utilizing either the UL or DL cellular resources. The author presented his results through simulations showing the case when DM mode is always selected regardless of the distance between the DUEs and the case in which CM mode is selected. It has been shown that the average sum rate is always highest for the proposal, investigating all the distances between the DUEs. The mobility of users is however, not considered by the author which can greatly affect the mode selection scheme.

We have observed that the mode selection for D2D communication has significant effect on the interference between DUEs and CUEs and efficient and dynamic selection can significantly improve the QoS to the network users. Moreover, the signaling overhead to the base stations can also be controlled through better selection of D2D communication mode. The literature has considered simple network scenarios where only few D2D pairs and cellular users are considered for analysis. Moreover, most of the work is done considering single tier network however, 5th generation networks are expected to be multitier with huge number of small cells and user densification. D2D communication is expected to be enabled in multitier heterogeneous networks and channel conditions and network dynamics will be a lot more complexer. Mode selection schemes in such complex and dense multitier D2D enabled Hetnets are required to be developed to fulfill requirements of 5G.

\subsection {Interference Mitigation through Power Control and RRM}
As discussed previously, the D2D communication can best pay off if it takes place in shared mode also known as underlay mode of communication. Underlay mode present complex challenges in terms of interference control between the DUEs and CUEs. Interference control is easier in dedicated mode or overlay mode but the spectral efficiency is quite lower. In this section, different interference mitigation techniques based on power control and radio resource allocation are discussed.

The interference caused by DUEs to the CUEs is most important. To control this interference, a transmission power control of the DUEs based technique is developed~\cite{yu2009performance}. The authors considered a system model with base station in center and circular coverage of radius \textit{R}. A single cellular user and one D2D pair operating inside the cell are sharing same frequency resource. D2D pair is confined to distance \textit{L} between them and D2D communication cannot take place beyond this distance \textit{L}. The interference models and SINR for cellular user and D2D pair has been constructed and power transmission of D2D pair is calculated for different distance form base station that causes only 3 dB degradation in the SINR of cellular user. The cellular communication can take place without errors if it faces 3 dB SINR degradation. The author collected results for both UL and DL frequency resource and found that SINR of D2D pair fluctuates from -30dB to -7dB in UL case and fluctuates between -20dB to -15dB in DL case. The problem with this simple power control technique is that the probability of D2D communication is very low because of low transmission power of the DUEs. A similar technique is presented in~\cite{janis2009interference} in which the power levels of DUEs is set by the base stations to achieve a target SINR but the performance benefits of this technique are not higher compared to ~\cite{yu2009performance}.

Interference mitigation is also done using radio resource allocation based on some criterion. The authors in ~\cite{wang2012distance} have used a distance-based resource sharing criterion (DRC) to allocate resource blocks to the D2D pairs and cellular user. The author considered a model with base station in center with circular coverage radius R. A D2D pair and a cellular user are considered inside the single cell coverage. The distance between D2D pair and cellular user \textit{L} is considered as the criterion for resource allocation. The D2D user and cellular user are using same frequency resources. The author assumed that GPS locations of all users are known to the base station and he assumed a minimum distance $L_{min}$ below which D2D communication cannot take place because of interference from cellular user. The author calculated SINR for D2D communication for different distances $L_{min}/R$ and the outage probability of D2D communication is plotted against distance \textit{L}. The results have shown that with distance-based resource sharing criterion (DRC), the outage probability of D2D communication can be significantly decreased thus enabling D2D communication in most of the cell area.

The authors in ~\cite{yu2011resource} have further improved the results by considering a better system model for 5G communication. The authors in ~\cite{yu2011resource} considered \textit{N} cellular users distributed uniformly in the circular cell coverage of radius \textit{R}. The author utilized location estimation algorithm and further decreased the outage probability of D2D communication by choosing the resource sharing of that cellular user that minimizes the outage probability of D2D communication. The interference models used by ~\cite{yu2011resource} are same as used by the [17] with the addition of location of cellular user taken into account. The results have shown that most of the cell area is operable for D2D communication and SINR of D2D user does not fall below the preset threshold. The authors in ~\cite{yu2011resource} also compared his results with ~\cite{wang2012distance} and shown the decrease in outage probability using his proposed algorithm. The advantage of these techniques is the reduced overhead to the base stations.

Another simple method to control interference form DUEs to CUEs is suggested in ~\cite{peng2009interference}. In this method, the base station calculates the tolerable interference levels when D2D uses the resource blocks (RB) in the UL and broadcast this information to the DUEs. The DUEs use this information to choose those RBs in the UL which cause minimal interference to the CUEs. The author presented performance improvement in terms of throughput of CUEs from 2.65 Mbps to 3.33 Mbps. However, this was achieved at the cost of throughput decrease of DUEs from 3.02 Mbps to 2.83 Mbps.

The problem of resource allocation to the DUEs is addressed in a different manner in ~\cite{zulhasnine2010efficient}. Authors have formulated the problem of resource allocation as Mixed Integer Non-Linear Programming (MINLP). MINLP is of very high complexity and its practical implementation is not feasible because the algorithm cannot be solved in 1 ms transmission time interval (TTI) considered in LTE (A).  In order to make the solution practical, a heuristic greedy algorithm is proposed ~\cite{tapia2009hspa}. The resource allocation is done based on the channel quality of the CUEs. If a CUEs are experiencing good channel quality, their resource is shared with the DUEs to keep their SINRs above certain thresholds. Extensive simulations have been performed by the author considering conventional allocation schemes like proportional Fair and Round Robin. It has been shown that the network throughput is substantially increased in this D2D enabled network. The author did not prove the optimality of this heuristic algorithm-based allocation.

Joint power and resource allocation-based techniques have also been developed to control the interference between DUEs and CUEs. In ~\cite{gu2011dynamic}, authors have presented a dynamic power control and resource allocation-based technique to mitigate the interference. The base station assigns the resources to the CUEs in a prioritized manner and remaining resources are allocated to the DUEs. Afterwards, if the requirement of DUEs is not met then base station determines the resources of CUEs that can be shared with DUEs. The interference among such DUEs and CUEs sharing resources is mitigated through dynamic power control. The power control is done by the base station which determines the transmission power of the DUEs based on channel quality measurements between DUEs and CUEs and between DUEs and base station in UL direction. Author compared his results with fractional power control-based technique and demonstrated a 5.7 dB increase in the SINR of CUEs and 2.77 dB increase in SINR of the DUEs. The power control management by base stations added to the significant overhead to its processing and this aspect is not discussed in the paper.

A joint resource allocation and power control technique based on column generation method to reduce to complexity of the problem has been proposed in ~\cite{phunchongharn2013resource}. Authors have considered a network with base station in the center and users distributed uniformly in the area of 50 m around it. The DUEs are randomly distributed in the area with distance between them as uniformly distributed from 0 to 25 m. The resource allocation problem objective is to reduce the interference caused to the cellular users and maintain the QoS of D2D users. One RB is shared with multiple D2D users to increase the spectral efficiency and the author has compared the results of his technique with the technique in which RB is shared with single D2D user. The D2D user calculates interference on all RBs and selects a RB that causes minimum interference to cellular users using same RB. If the interference caused is under certain threshold and access constraints are met, it reuses the RB otherwise it searches for other RBs. The objective of the reuse of RB is to reduce the transmission time interval which in turn translates to maximizing the spectral efficiency. The author has presented his results showing that with the little increase in the transmission power of D2D links, the transmission time interval can be significantly reduced thus increasing spectral efficiency. The technique is centralized and will be running on the central base station therefore it is suited for smaller and medium sized networks in which traffic demands do not change very fast.

The mutual interference between DUEs and CUEs is solved through Fractional Frequency reuse (FFR) approach ~\cite{chae2011radio}. The frequency band available to the base station is divided into four sub frequency bands (f1, f2, f3, f4). The inner region of his base station uses the sub-band f1 and while other sub-bands are utilized by outer regions. The D2D pairs also exploit this division and utilize the frequency bands of other regions when located in inner region. For example, if a D2D pair is located in region of band f1, it will reuse the resources being used in region with band f2 or f3. In this way, the interference caused to the cellular user and other D2D users will be under control and QoS for all network users is easily met. The problem with FFR approach is that the bandwidth is not efficiently utilized and it is dependent on accurate location estimation. Error in location estimation will result in very bad performance.

Graph theory has also been employed to solve the problem of interference between DUEs and CUEs in cellular networks. An interference aware graph theory-based resource allocation scheme is presented in~\cite{zhang2013interference}. Sum rate maximization of DUEs and CUEs through resource allocation is the objective of this scheme. The interference among DUEs and CUEs are represented in the form of graph. An interference graph is constructed first based on the network topology. The graph has three main characteristics; 1) the link attribute which tells whether the vertex in a graph is for DUE or CUE; 2) SINR values for each resource block; 3) attribute representing allocation of RB to the individual vertexes. The optimal solution for allocation of resources is based on exhaustive search that tries all possible combination of allocation possibilities therefore author implemented sub optimal solution to reduce the complexity of the scheme. The results show that the performance of sub optimal solution is almost same as of optimal solution and it significantly maximizes the sum rate and is spectrally efficient compared to greedy orthogonal sharing scheme.

The interference problem is solved through game theory in ~\cite{wang2013joint}. Authors have solved two optimization problems, one related to the resource sharing and other relating to the optimal transmission power selection. This optimization was solved using Stackelberg game. The CUEs are made as the leaders and DUEs as the followers and leader is made as the owner of the radio resources. The followers are charged a certain fee if they use the same resources. The utility function based on the throughput of the leader is defined and first optimization problems requires setting up of a price to maximize this utility function. The second optimization problem is to set transmission power of the DUEs (followers) to maximize their utility function. A joint resource scheduling and power allocation is done afterwards to fairly distribute resources among the DUEs and CUEs. The author has proved in his results that the throughput of DUEs increases with increase in CUEs admitted to the network, because more resources will be available to be shared with DUEs.

The interference mitigation and resource allocation schemes developed by researchers discussed till now considered simple network scenarios and did not considered multitier heterogeneous network models. The network model considered have few D2D pairs and CUEs with only one macro base station in center which is not practical in nature therefore, researchers identified this shortcoming and have done analysis using concepts of stochastic geometry and multitier heterogeneous networks. The complexity of interference in multitier Hetnets is much more than single tier network because each tier will cause interference to the other tier (see figure \ref{fig:Hetnet2}).

A successive interference cancellation (SIC) scheme for stochastic geometry-based network model has been presented in ~\cite{ma2015performance}. The author employed concepts of stochastic geometry and considered a network model with DUEs and CUEs distributed as per Homogeneous Poisson Point Process (PPP) and base stations distributed using Stationary Point Process. The base stations are assumed to have infinite SIC capability while D2D receivers have finite SIC capability. The author has presented the stochastic equivalence of the interference, by which a two-tier network (Macro and D2D tier) can be represented by a single tier interference. The successful transmission probabilities of CUEs and DUEs are calculated for equivalent model and are validated by simulations and analytical results.

A network assisted interference mitigation scheme is presented in ~\cite{tsai2012intelligent} in a two-tier heterogeneous mobile network with macro and femto base stations. The UL of OFDMA based network model is considered and it has one macro base station and cluster of femto cells deployed inside the houses. The D2D pairs are located in the coverage of the femto cells and share the spectrum of the cellular users. The Carrier to interference plus noise ratio (CINR) for different sub carriers of OFDMA system are calculated for macro BS, femto BS and D2D users. Based on these measurement, macro base station calculates the tolerable power levels and broadcast this information to the D2D users to keep the interference levels below predefined thresholds. The minimum transmission power of DUES is also calculated to keep their SINR above the predefined $SINR_{DUE}$ threshold. The scheme is reliable as it ensures the link reliability of femto and macro cell users as well as the D2D users in the network however, this scheme requires broadcast of information to the D2D users without which this scheme renders useless.

The interference caused by DUEs to the macro and femto cells is handled through Stackelberg game in ~\cite{he2014resource}. UEs of the macro and femto cells are taken as leaders and DUEs as the followers. The leaders own the radio resources and charge fees to the DUEs for using these resources. The author assumed that the macro BS and Femto BS use dedicated channels however, a more realistic scenario must be considered in which macro and small cells share channels with each other.
\subsection {Summary of Conventional Interference Mitigation and RRM techniques}
From the papers we reviewed in this survey, we observed that most of the power control techniques are used to solve interference caused by DUEs to the CUEs. The power techniques are strongly dependent on the distance of the DUEs from CUEs if DL resource is shared and from base station if UL resource is shared with the DUEs. The decrease in transmission power of DUEs due to these distances can also affect the QoS of DUEs if the distance between them is larger. Therefore, power control techniques become useless in certain cases. In such cases, RRM techniques are used to control the interference. The resource allocation can either be done by the base stations in full control mode in which the base station needs to know the CSI of all the links involved in communication or it can be done the DUEs themselves in loose control mode which also has the advantage of reduced signaling overhead. Loose control however is not desired by the network operators because they will lose control over the network management.

Joint power control and radio resource allocation schemes are better solution to control interference among DUEs and CUEs. Other interference control techniques include massive MIMO, beamforming and Interference alignment techniques. Most of the papers considered scenario 1C (see figure \ref{fig:D2DScenarios}) in which both D2D users are in coverage of the same cell and they share resource of one CUE with only one D2D pair. Only few papers assumed multiple CUEs and DUEs in their network models. The mobility of users is not considered by the papers which will significantly change the interference scenario and require more sophisticated techniques to handle interference.

\section{Artificial intelligence and machine learning based interference mitigation and rrm techniques}
\begin{figure}
	\centering
	\includegraphics[width=1\linewidth]{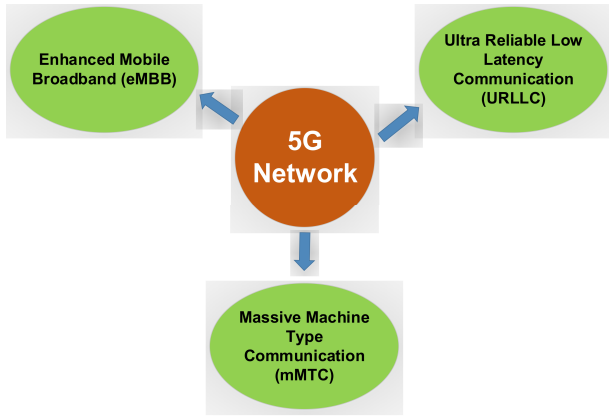}
	\caption{Main Services of 5G}
	\label{fig:5GReq}
\end{figure}

Currently 4G network is providing the seamless Internet Protocol (IP) based connectivity to all mobile devices. It took almost 30 years to successfully transform the conventional telephone based mobile communication to fully digital IP based communications. Internet of things has given rise to more connected devices and throughput requirements have increased due to high quality video streaming and entertainment applications. 5th generation networks are envisioned to meet these growing requirements as described earlier in this paper however, networks need to revolutionized with cutting edge technologies to meet these demands. The three main services provided by 5G (see figure \ref{fig:5GReq}), namely Enhanced Mobile Broadband (eMBB), Ultra Reliable Low Latency Services (URLLC) and Massive Machine Type Communications (mMTC), will be enabled by technologies like cell densification through small cells and MIMO but these technologies are cost ineffective. A cost-effective technology for enabling these services is to exploit Artificial Intelligence (AI) for network functions like Radio Resource Management (RRM), Mobility Management (MM) and Orchestration (MANO) and Service Provision Management (SPM).

AI is the science of making the machines intelligent just like humans. It has been effectively used to solve diverse problems of nature and has been applied to communication problems as well for optimized results. AI falls into two categories, one in which the machine has predefined options of action to select from and choose the best action among them. The second category of AI is the one in which, the machine interacts (e.g. sense, mine, reason and predict) with the environment and then take actions for optimal results. In communication networks, the second category of AI is applied as the channel conditions and network parameters keep on changing and devices or base stations need to sense these changes continually to taker better actions. Owing to the effectiveness of AI, the complex interference and RRM problems have also been addressed by different researchers. Different AI based algorithms like Genetic Algorithms (GA), Ant Colony Optimization (ACO) and Reinforcement Learning (RL) have been employed to solve interference problem between DUEs and CUEs.

Genetic Algorithm is inspired by the process of Natural Selection and they generate high quality solutions to optimization problems. GA is based on Swarm Intelligence and reaches optimal solution with faster speeds and lower complexity. A Genetic Algorithm based technique is presented in ~\cite{yang2014ga} in which resource sharing between DUEs and CUEs is done through GA. The authors considered a network scenario with 30 CUEs and 30 DUEs in a single cell and 50 Resource blocks to be shared among them. Number of D2D users, number of CUEs and number of resource blocks available are considered for coding required for genetic algorithm. Afterwards, optimization is performed to increase system throughput. Results are compared with Exhaustive search algorithm which is very computational extensive, greedy heuristic algorithm, network with no DUEs and network with D2D communication enabled and using orthogonal resources. A throughput gains of 30 Mbps over greedy Heuristic algorithm is achieved while throughput achieved is 5 Mbps less than exhaustive search algorithm. Computational requirement analysis has also been done and comparison is made with Exhaustive Search algorithm. The network scenario considered however is single cell based which is not practical for 5th generation networks and user mobilities are not considered which will significantly affect the results.

ACO is also a swarm intelligence-based algorithm based on probabilistic techniques to solve computational problems which can be reduced to finding optimal paths through graphs. It is inspired by the behavior of ants who move in search of food between their colonies. ACO algorithm has been employed in ~\cite{liotou2014ant} to solve the interference problem in D2D enabled cellular network. Graph coloring is used for mapping the interference among D2D users using Interference Level Indicator (ILI) term and numerical values from 1-15 are used to quantify the interference. After that, Ant colony optimization (ACO) algorithm is used to allocate Resource Blocks among D2D pairs to increase spectrum reuse. A single cell network scenario with D2D users, Cellular users and single base station is considered. Target outage probability is selected by choosing minimum outage probability threshold and minimum SINR threshold. eNB calculates the interference levels among the D2D users using channel information and path loss and graph model is created representing the interference among D2D users. Iterative approach is used to choose the optimum number of D2D links that gives the outage probability lesser than the predefined threshold. Afterwards, ACO is used by minimizing the sum of weights (mutual interference). Network parameters as per LTE standard have been chosen and convergence of ACO algorithm in a graph representing the cost measurement and number of evaluation runs is presented. The results are compared with exhaustive search algorithm which is computationally extensive and comparable spectrum efficiency with quiet lower computational requirements is achieved. Computational requirements comparison of ACO with exhaustive search algorithm has also been done and significant improvements are observed. The author recommended to include mobility of D2D users in network model as future work.

Reinforcement learning (RL) has been greatly employed in solving the network problems in 5G heterogeneous networks. Reinforcement leaning is a machine learning based technique that does not require any model to predict the future actions or draw inferences. \textit{Q}-Learning, a sub part of reinforcement learning, has been used by many researchers to solve resource allocation, cell association and interference mitigation problems. It is a model free learning technique in which the learning agent tries to maximize its reward by taking immediate actions. Due to the uncertainty of the 5G mobile networks due to changing network conditions, channel fading and user mobilities, network cannot be modeled therefore traditional model-based learning schemes cannot be employed. Therefore, \textit{Q}-Learning has proved to be a powerful tool to solve network problems. \textit{Q}-Learning includes four parameters namely action \textit{a}, state \textit{s}, transition probability form one state to other state $P_{s,s^\prime}$ and reward $r_{s, a}$. The state is the internal phenomenon of each agent while reward reflects the quality of action taken by the agent. The objective of the \textit{Q}-Learning is to determine the optimal policy ${\pi }_{s}^{\ast }$ to choose the actions that give the maximum reward. The process works as follows; the agent chooses an action at time \textit{t} in some state \textit{s} and measure the reward \textit{r}. The agent records the reward and move to next state $s^\prime$ to choose the next action. The \textit{Q} value for each action is measured and recorded in \textit{Q} matrix of size state x action according to following equation:'
\begin{equation} \label{eq:15}
\begin{aligned}
 Q(s,a) = (1-\alpha) Q(s,a) + \alpha [r(s,a)\\ +\gamma\max\limits_{b \in A}Q(s^\prime ,b)]
 \end{aligned}
\end{equation}

Where $\alpha$ is the learning rate and $\gamma$ is the discount factor. An appropriate action is assigned a positive reward and hence gets a high \textit{Q}-value while inappropriate action is punished and gets lower \textit{Q}-value. It has been proved in ~\cite{watkins1992q} that the update rule of \textit{Q}-values in a two dimensional look up table converges to optimal \textit{Q}-value when state and actions are visited infinitely often. The learning comprises of two stages namely exploration and exploitation. In exploration phase, the agent explores all states and actions and record the \textit{Q}-values while in exploitation phase, only those actions are chosen whose \textit{Q}-values are higher.

A \textit{Q}-Learning based resource allocation scheme has been developed in ~\cite{luo2014dynamic}. A simple network scenario with 2 DUEs and 2 CUEs is considered. 2 channels [Ch1, Ch2] and 3 power states [P1, P2, P3] making six combinations are allocated to the DUEs. User locations, user channels and user arrivals are taken as input to the algorithm and the objective is to maximize the system capacity calculated through Shannon capacity formula. Author has considered just 2 CUEs and 2 DUEs in a single cell network model which is not practical and inputs for decision making are also too simple. System capacity through \textit{Q}-learning is compared with random resource allocation and maximum power allocation and performance gains are presented.

A multicell network model is considered in ~\cite{alqerm2016cooperative} for mitigating interference among DUEs and CUEs and intercell interference through \textit{Q}-learning based resource allocation scheme. The network model considered contained a macro cell, cluster of femto cells with one UE in each femto cell and D2D users distributed in the coverage of macro cell. The authors considered N resource blocks to be allocated with P= (1,2,....,P) power levels and M= (1,2,....,M) modulation indexes chosen intelligently to control the interference and maximize D2D users throughput and spectral efficiency. SINR of D2D receiver is formulated and that of Macro users and state-action pairs were made. The \textit{Q}-values for each pair is determined for particular allocated resource and \textit{Q}-table is made by the devices and it is shared among all users of the network to find the optimal resource sharing policy. The constraints considered include SINR threshold of macro user greater than predefined threshold, only one resource block to be used by each user with one power level and modulation index and a binary decision variable which outputs 1 with transmitter selecting resource block N, power level P and modulation index M otherwise it will be zero.

Authors have done resource allocation in two phases. In exploration phase, users select different actions and measure rewards to explore best rewarded action and in exploitation phase, the action with maximum reward (high \textit{Q}-value) is selected. Author defined exploration rate $\epsilon$ which is higher in start to find highest Q-value and learning rate $\alpha$ which is faster when higher \textit{Q}-value is not found and when it is found, $\alpha$ becomes lower. All users (agents) maximize their local \textit{Q}-values and global \textit{Q}-value is decomposed into the linear combination of local \textit{Q}-values thus if each agent maximizes its \textit{Q}-value, global \textit{Q}-value is maximized. The results of his scheme are presented showing spectral efficiency and throughput improvements in comparison with joint-Resource Allocation and Link Adaptation (RALA) scheme, Matching RM and Down SA schemes. The drawback in this scheme is that the network model considered has one UE associated with each femto cell and only one macro cell is considered. The effect of macro to macro cell interference is not addressed in this paper. Moreover, the DUEs and Femto cell users need to share their \textit{Q}-tables to find the best allocation scheme which is itself an overhead for the network.

Similar to ~\cite{alqerm2016cooperative}, authors in ~\cite{khan2017throughput} have presented a cooperative reinforcement learning technique for allocating RBs and power level to D2D users underlaying cellular users. A single tier netowrk model with macro base stations is considered and D2D and cellular users operate under the coverage of macro base stations. The authors have included cooperation between the learnign agents which are D2D users in whihc they share their value functions to jointly increase the overall throughput of the system. A comparison of increase in system throughput and fairness of allocation is made with distributed reinforcement learning based technique with no coperation~\cite{luo2014dynamic}~\cite{nie2016q} and random allocation technique. The drawback of this paper is that it considered a single tier network however, 5G network is expected to be multi tier heterogeneous network.   

An expected \textit{Q}-learning technique to find optimal resource allocation scheme for optimizing user’s data rates and spectrum usage in a decoupled LTE-U network is presented in ~\cite{hu2017expected}. A game theoretic model incorporating user association, spectrum allocation and load balancing is considered for resource allocation among the LTE-U and WiFi users. The network model considered has a macro cell base station in the center and $N_s$ small cell base stations, $N_u$ LTE-U users, W wireless access points (WAP) and $N_w$ WiFi users uniformly distributed. Macro cell users utilize the licensed band in uplink and downlink while LTE-U and WiFi users utilize unlicensed spectrum for communication. WAP utilizes CSMA/CA (carrier sense multiple access with collision avoidance) protocol for spectrum usage. The resources to be allocated to the users consist of uplink and downlink bands in licensed spectrum and time slots for LTE-U and WiFi users in unlicensed band. Authors considered logarithmic function to compute utility functions which make resource allocation fair among the user with different data rates. Authors have developed a \textit{Q}-learning based allocation scheme where each base station allocates resources based on $\epsilon$-greedy mechanism (exploration) and measure \textit{Q}-values and update its state while broadcasting its \textit{Q}-value to the other base stations. The other base stations will use this information to determine its resource allocation scheme and share it with others. In this way, a global maximum \textit{Q}-value is achieved to find the optimal resource allocation for required data rates. If the data rate of some users is below the required data rate, then it will keep sending requests to base stations for connection to get better data rates. In this way, algorithm finds a mixed strategy Nash Equilibrium to optimize data rates of all the users. The authors presented their results showing increase of 12.7 \% and 51.1 \% in sum rates (UL+DL) compared to traditional \textit{Q}-learning and LTE-U nearest neighbor algorithm respectively. Authors have also shown that is algorithm takes 19 \% less time to converge to the optimal solution. The drawback of this paper is the requirement to share learnt information with neighboring agents which adds to the overhead.

An autonomous \textit{Q}-Learning based technique is presented in ~\cite{asheralieva2016autonomous} in which D2D users autonomously select resource blocks and power levels in a distributive and decentralized manner to mitigate interference between DUEs and CUEs. The authors have considered joint operation of cellular users and D2D users in a heterogeneous cellular network with multiple Base Stations and D2D users. Authors have done analysis of network in two scenarios (i) when orthogonal resource is shared among cellular and D2D users (ii) and when resource is shared among them. The goal of each D2D pair is to jointly select the wireless channel and power level to maximize its reward which is the difference between throughput and cost of power consumption with the constraint of having a minimum SINR requirement. A cooperative game-based approach is used with multiple D2D users as players who learn their best strategies based on locally observed information and developed a fully autonomous multi agent \textit{Q}-learning algorithm converging to a mixed strategy Nash Equilibrium (NE). Authors considered a heterogeneous network as per 3GPP LTE-A standard with 3 Base Stations (macro, micro and femto each) with 100 cellular and 100 D2D users distributed randomly and considered standard time division duplexing (TDD) scheme. \textit{K} orthogonal resource blocks and \textit{J} power levels are considered for allocation to the D2D users for both scenarios. The reward is measured as sum of instantaneous rewards over infinite time interval as time for which user stays in network is unknown. Therefore, a discounting factor $\gamma$ is introduced to avoid the infinite sum problem.

The authors have made a matrix comprising of all possibilities of channel and power allocation and aims to find the best pair of channel and power level that maximizes the reward. Each D2D pair selects an action containing channel-power level pair to maximize its own reward and does not know about the actions of other D2D pairs. The selection of channel and power level is done at particular instant without knowledge of previous instances therefore making it a Markov Decision Process (MDP). A Multi-Agent Q-learning based scheme is developed for such MDP in which each D2D pair determines its optimal strategy for action selection. The strategy is made to maximize the value state function \textit{V} defined by authors that maximizes the expected value of the reward function and achieves Nash Equilibrium. More than one strategy can exist for each learning agent to achieve Nash Equilibrium therefore authors select the state function with maximum value. For autonomous selection of actions by D2D users without knowledge of other D2D pair actions, each D2D pair estimates the beliefs about other player’s strategies. Reference points for the beliefs and strategies were chosen and were continuously updated based on previous time slot measurements of beliefs and strategies. Based on the estimation of belief, D2D pairs autonomously select actions to maximize their rewards.

In order to overcome the challenge of exploitation/exploration tradeoff of \textit{Q}-learning, author considered $\epsilon$-greedy selection scheme and all actions were weighted according to their action values (rewards) so that actions with higher probability are selected. Authors have made use of Boltzmann Gibbs distribution for action selection and presented results showing effect of temperature $T_B$ (used in Boltzmann Gibbs Distribution) on the convergence of autonomous channel and power level section (ACS). Authors have also presented the effect of increase in D2D users on the throughput of both cellular and D2D users and compared his results with $\epsilon$-greedy \textit{Q}-learning based action selection scheme, uniform random selection scheme, parallel fictitious play (FP), parallel best response dynamics (BRD) based action selection scheme where action is selected according to CSI information and previous actions and optimal centralized strategy (OCS) where action is selected according to global CSI information. Authors have also presented results showing $SINR_{min}$ value selection on the throughput of cellular and D2D users and shown significant improvements in the network.

Random Forest alogrithm has been employed by authors in~\cite{imtiaz2016learning} to allocate resources in a Time Division Duplexing (TDD) based Cloud Radio Access Network (CRAN). Authors have considered a CRAN system with Remote Radio Heads (RRH) deployed to provide Line of Sight (LOS) communication to a large number of users. The overhead of gathering instant CSI for high mobility users is tremendously large in ultra dense network therefore, the proposed technique exploits the position estimates of high mobility users which are somewhat predictalbe~\cite{lu2013approaching} to allocate the resources to the users. This waves off the requirement of CSI for resource allocation tasks. The robustness of the proposed  scheme is tested by using accuarte position estimates in training dataset and inaccurate estimated in test datasets for random forest. The system throughput is calculated afterwards and comapared with the CSI based resource allocation scheme. Significant perfornace gains are achieved by proposed technique with overheads of 2.5\% compared to 19\% overhead in CSI based technique. The assumptions of LOS communication and requirement of accurate position estimates for training dataset limits the performance of this scheme.    

An energy efficient power allocation scheme is presented in ~\cite{alqerm2017energy} in which power is allocated to the users using enhanced online learning. The allocation is done is non cooperative manner to maximize the energy efficiency of the network. The devices select the power levels in distributed and autonomous manner based on intuition about the other devices power selection strategies. The power selection strategy is determined by the devices using \textit{Q}-learning algorithm with reduced states to increase the convergence time of the algorithm. The authors have considered a two tier heterogeneous network model where the first tier comprises of the macro base stations and second tier consists of femto or pico cells and D2D users. The downlink transmission model is considered for power allocation with the aim of maximizing energy efficiency. Authors have proved that the energy efficiency can be significantly increased and spectral efficiency can be enhanced using the \textit{Q}-learning based enhanced online structure and convergence times can be reduced using intuition based power allocation to the devices.

Heterogeneous Cloud Radio Access Netowrk (H-CRAN) have become a focus in 5G Networks to leverage the benefits of both heterogeneous and CRAN advantages. Therefore, authors in~\cite{alqerm2017enhanced} have presented an enhanced machine learning scheme for energy efficient resource allocation in 5G H-CRAN. The network model considered has macro base stations and Remote Radio Heads (RRH) which serve two types of users, one with high QoS requirements and other with low QoS requirements. The Q-learning methodology is employed to exploit the low power RRHs for interference mitigation between macro tier and RRH tier while meeting QoS requirements of the cellualr users and maximizing energy efficiency. The availble resource blocks (RB) are divided into two sets of RBs, one for users with high QoS requirements located at cell edges and served by RRHs and other for sharing with RRH users and macro cell users located in the center and having low QoS requirements. The learning methodology is employed separately for both sets of RBs to reduce the convergence time of the algorithm. The resource llocation is done centrally by centralized Baseband Unit (BBU). All users report their channel state information and path losses to thier serivng base stations and RRHs, which is then sent to the centralized controller. The controller exploits this information to learn the environment and allocate power and RBs to the users in order to maximize energy efficiency and maintain QoS requirements of the served users. Significant performance gains interms of energy efficiency, spectral efficiency and data rates are achieved through this centralized online learning scheme. The limitaiton of this scheme is that if the central controller fails tooperate, the whole netowrk will go downand wuill not function.  

Therefore, authors in ~\cite{alqerm2018sophisticated} have extenteded the work in~\cite{alqerm2017enhanced} to include decentralized resource allocation in the network in addition to centralized resource allocation by the BBU. In decentralized resource allocation, the macro base stations allocate resources to the RRHs and cellualr users as they have all chanel state infromation and path losses form the users and RRH operatin gunder their coverage. The learning is implemented in a distributed manner in all macro base stations in whihc they learn a common strategy $\pi$ to allocate RBs and power level to the users to maximize the system energy efficiency while meeting QoS requirements. The authors have alos implemented both cnetralized and decentralized techniques on Software Defined Radio (SDR) plaatforms to practically test the performance of the schmemes. The hardware setup comprises of GNU radio, USRP N210 from Ettus Research ~\cite{SDRForum} and two dell servers for base band processing. The numerical and practical results have shown considerable increase in energy efficiency and spectral effciency of the system while providing higher bit rates and low Bit Error Rates (BER) in the system.

Authors in~\cite{chattopadhyay2016location} have used the concept of stochastic learning for opportunistic bandwidth sharing between static and mobie users in a cellular network. The channel conditions keep on changing for the mobile users at a faster rate compared to static users and mobile users keep on changing thier serving base stations. It is quite challenging to provide higher data rates to these moving mobile users. Therefore, authors in ~\cite{chattopadhyay2016location} have proposed a location dependent bandwidth sharing and formulated the problem as a long run average reward Markov Decision Process (MDP). The reward function of such an MDP either changes with time or not known at all. To overcome these problems, a time scale stochastic approximation based learning algorithm is proposed. The authors have considered a multiple Macro Base station based netowrk model with static and mobile cellular users. The mobile users are moving along a line and keeps on changing their serving base station after fixed time slot $\sigma$ while moving at some velocity $v$. Authors have presented learning model for both constrained and unconstrained objective functions and formulated states, actions, state transitions and rewards for this MDP. The optimality and convergence of the learning algorithms have been proved and fairness of bandwidth sharing has been presented in the paper. It has been shown that significant improvements can be achieved through this stochastic learning model for mobile users in the cellular network. However, the authors have limited their analysis to the users moving with fixed velocity while in practical networks the users move with different velocities.  
 
Thus far we have seen that AI techniques have proved to be a powerful tool to address interference and resource allocation problems in 5th generation networks. The learning capability when put into the devices will distribute the processing load of the base stations and decision process will become decentralized. This decentralized decision will ease off base stations for performing other network functions like cell association, mobility management and other control tasks etc. Moreover, network performance can also be increased as depicted in the papers discussed above. Though genetic algorithm and ACO algorithm has been employed in resource allocation and interference mitigation problem, Reinforcement learning (Q-Learning) has proved to be a better solution to these problem because of its model free learning capability. The network scenarios considered in literature to solve RRM problem include heterogeneous networks however, user mobilities are not considered which can significantly change the network parameters and hence the solutions need to be tailored. Moreover, the out of coverage D2D communication scenarios are not considered in developing these learning algorithms.

\begin{figure*}
	\centering
	\includegraphics[width=1\linewidth]{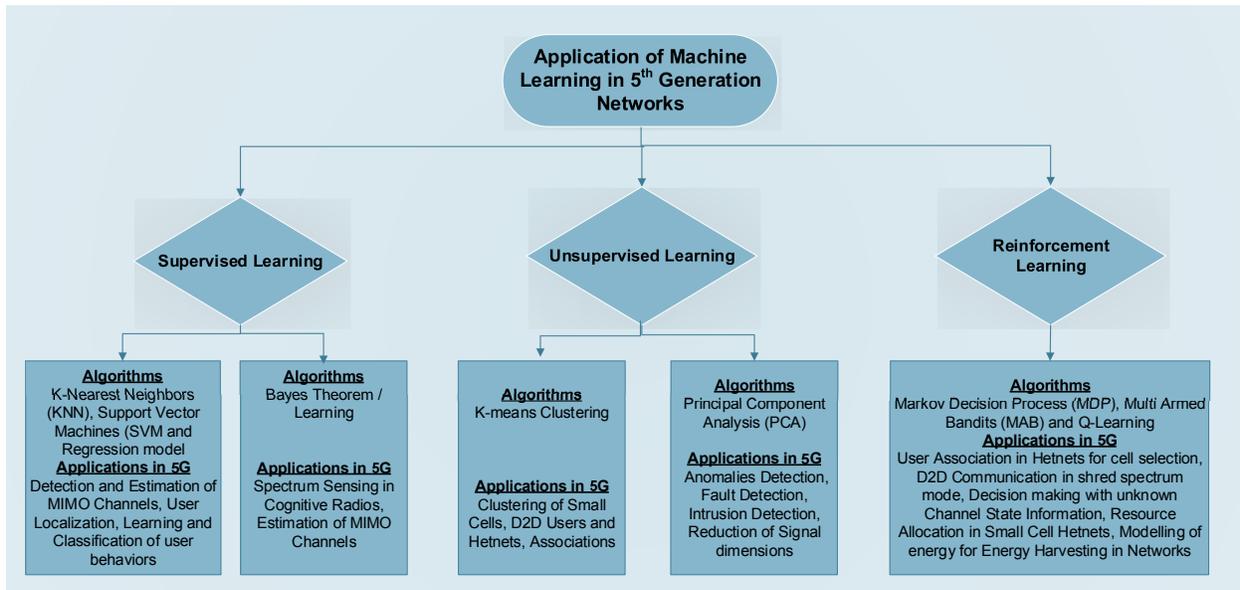}
	\caption{Machine Learning in 5G}
	\label{fig:ML}
\end{figure*}
The Artificial intelligence and Machine Learning techniques cannot be denied when it comes to solve 5G network problems. Other than RRM and interference mitigation, AI can also be employed in Software Defined Networking (SDN) and Network Function Virtualization (NFV), Self-Healing Networks, Intrusion Detection in network, channel estimation/detection and spectrum sensing and detection in Cognitive Radios (CR). Learning algorithms like Bayesian learning, K Nearest Neighbor (KNN), K-means Clustering, Support Vector Machines (SVM), Principle Component Analysis (PCA) and Markov Decision Process (MDP) plays a vital role in solving different network problems and enable envisioned 5G requirements ~\cite{alliance20155g}. The employment of Artificial Intelligence and Machine Learning techniques addressing different problems of 5G networks is depicted in Figure \ref{fig:ML}.

\section {Future research directions and challenges}
D2D communication has huge performance benefits including low latency communication, high data rates because of close proximity communications, high spectral efficiency and high energy efficiency. These performance benefits can enable the requirements laid down by 3GPP and NGMN alliance for 5th generation network. The achievement of these performance benefits requires efficient Radio Resource allocation and interference mitigation techniques to be developed.
\subsection{Mode Selection Schemes}
The interference mitigation through mode selection scheme needs to be fully dynamic and adopt to the changing network conditions. The heterogeneity in the network will create complex interferences therefore, dynamistic mode selection will significantly impact the network performance. Moreover, user mobilities need to be considered in network models because it affects the network performance and plays important role in dynamic mode selection. Interference mitigation through resource allocation needs to be dynamic too where the network adapts to the changing network and determine allocation schemes in dynamic manner. This is not only increase the network performance in terms of throughput but will also increase the spectral efficiency without compromising the cellular user’s performance. Most of the literature considered either UL and DL link shared by DUEs with the CUEs however sharing of both UL and DL resources in dynamic manner will further increase the network performance.
\subsection{Interference Mitigation in Heterogeneous Networks}
5th generation networks will have multiple cells including Macro, Micro, Pico and Femto cells and the number of small cells will significantly increase to support huge active users in the network. D2D communication in these multi cell heterogeneous networks will make interference mitigation quite complexer therefore conventional interference mitigation scheme will add significant overhead to the base stations processing because of larger number of users. The decision making requires huge number of calculations to be performed prior to allocating radio resources to the users for communication. Owing to the complexity of the problem, artificial intelligence-based techniques have been employed in literature. GA and ACO algorithms have been employed for RRM in single cell network scenarios however, employment of these algorithms in multicell heterogeneous networks needs to be explored. RL has been greatly employed and proved very useful in RRM and interference mitigation algorithms in 5G networks. The multicell heterogeneous networks have been considered in literature and \textit{Q}-Learning based schemes have been developed for RRM in both centralized and decentralized manner. Most of the literature has considered sharing of learnt information by the DUEs with other DUEs and base stations for efficient RRM however, the mechanism to share this information has not been discussed. Autonomous decision making by DUEs has also been explored in multicell network however, single macro base station is considered in the study. The practical network will have multiple macro cells and small cells and such network has not been considered in literature for RRM through RRM technique. Moreover, user mobilities have also not been considered in the network models while developing RRM and interference mitigation techniques.
\subsection{New D2D Scenarios and D2D Communication in mmWave and Unlicensed Bands}
D2D communication scenarios have been laid down by 3GPP as shown in Fig. 3. Almost, all of the literature has considered scenario 1C in which both D2D communicating devices are located in the coverage of same cell. The other scenarios of D2D communication (see Fig. 3.) needs to be investigated for example D2D communication when both devices lie in the coverage of different cells. In addition to it, D2D communication in mmWave band needs to be exploited. This has several advantages as mmWave has relatively higher bandwidth and can provide much higher data rates and smaller distance between DUEs can ensure efficient communication over mmWave bands. The usage of mmWave bands for D2D and other bands for cellular communication will not introduce any interference and resource allocation will be quite easier for the base stations. D2D communication in unlicensed bands can also be exploited because it can ensure efficient cellular communication however, interference mitigation for D2D communication in unlicensed bands will be quite complexer because of no control of network over these bands.

\section{Conclusion}
D2D communication underlaying cellular network can provide significant performance improvements in terms of throughput and spectral efficiency however, with these performance benefits there are several challenges related to management of DUEs, interference mitigation and allocation of radio resources to the DUEs. Interference mitigation and RRM for D2D pairs is an active research area and new techniques are being researched for efficient direct communication. Artificial intelligence has been greatly exploited to solve these complex interferences in 5th generation multicell heterogeneous networks. The major weakness in the research is the consideration of single macro cell based network models , lack of user mobilities in network models and lack of realistic network scenarios with densely deployed small cells. 5G networks are envisioned to be supporting huge density of users therefore network models needs to be more practical as per 5G requirements. Moreover, the true extent of artificial intelligence needs to be exploited in 5G heterogenous networks to meet the demands of increasing users.





%\bibliographystyle{IEEEtran}
\bibliography{references}{}
\bibliographystyle{ieeetr}

\end{document}